\documentclass[10pt,twocolumn,letterpaper]{article}

\usepackage{cvpr}
\usepackage{times}
\usepackage{epsfig}
\usepackage{graphicx}
\usepackage{amsmath}
\usepackage{amssymb}
\usepackage{framed,multirow}
\usepackage[ruled,algonl,lined]{algorithm2e}
\usepackage{tabularx}

\usepackage[pagebackref=true,breaklinks=true,letterpaper=true,colorlinks,bookmarks=false]{hyperref}

\def\cvprPaperID{***} % *** Enter the CVPR Paper ID here

\cvprfinaltrue
\ifcvprfinal\fi
\begin{document}

%%%%%%%%% TITLE
\title{Real-time visual tracking by deep reinforced decision making}

\author{Janghoon Choi\\
Department of ECE, ASRI, Seoul National University\\
{\tt\small ultio791@snu.ac.kr}
\and
Junseok Kwon\\
School of CSE, Chung-Ang University\\
{\tt\small jskwon@cau.ac.kr}
\and
Kyoung Mu Lee\\
Department of ECE, ASRI, Seoul National University\\
{\tt\small kyoungmu@snu.ac.kr}
}

%\author{Janghoon Choi$^{1,}$, Junseok Kwon$^{2,}$, and Kyoung Mu Lee$^{1}$\\\\
%	$^{1}$ASRI, Seoul National University, $^{2}$School of CSE, Chung-Ang University\\
%}

\maketitle
%\thispagestyle{empty}

%%%%%%%%% ABSTRACT
\begin{abstract}
One of the major challenges of model-free visual tracking problem has been the difficulty originating from the unpredictable and drastic changes in the appearance of objects we target to track. Existing methods tackle this problem by updating the appearance model on-line in order to adapt to the changes in the appearance. Despite the success of these methods however, inaccurate and erroneous updates of the appearance model result in a tracker drift. In this paper, we introduce a novel real-time visual tracking algorithm based on a template selection strategy constructed by deep reinforcement learning methods. The tracking algorithm utilizes this strategy to choose the appropriate template for tracking a given frame. The template selection strategy is self-learned by utilizing a simple policy gradient method on numerous training episodes randomly generated from a tracking benchmark dataset. Our proposed reinforcement learning framework is generally applicable to other confidence map based tracking algorithms. The experiment shows that our tracking algorithm runs in real-time speed of 43 fps and the proposed policy network effectively decides the appropriate template for successful visual tracking.
\end{abstract}

%%%%%%%%% BODY TEXT
\section{Introduction}

\begin{figure}[t]
	\centering{\includegraphics[width=0.90\linewidth]{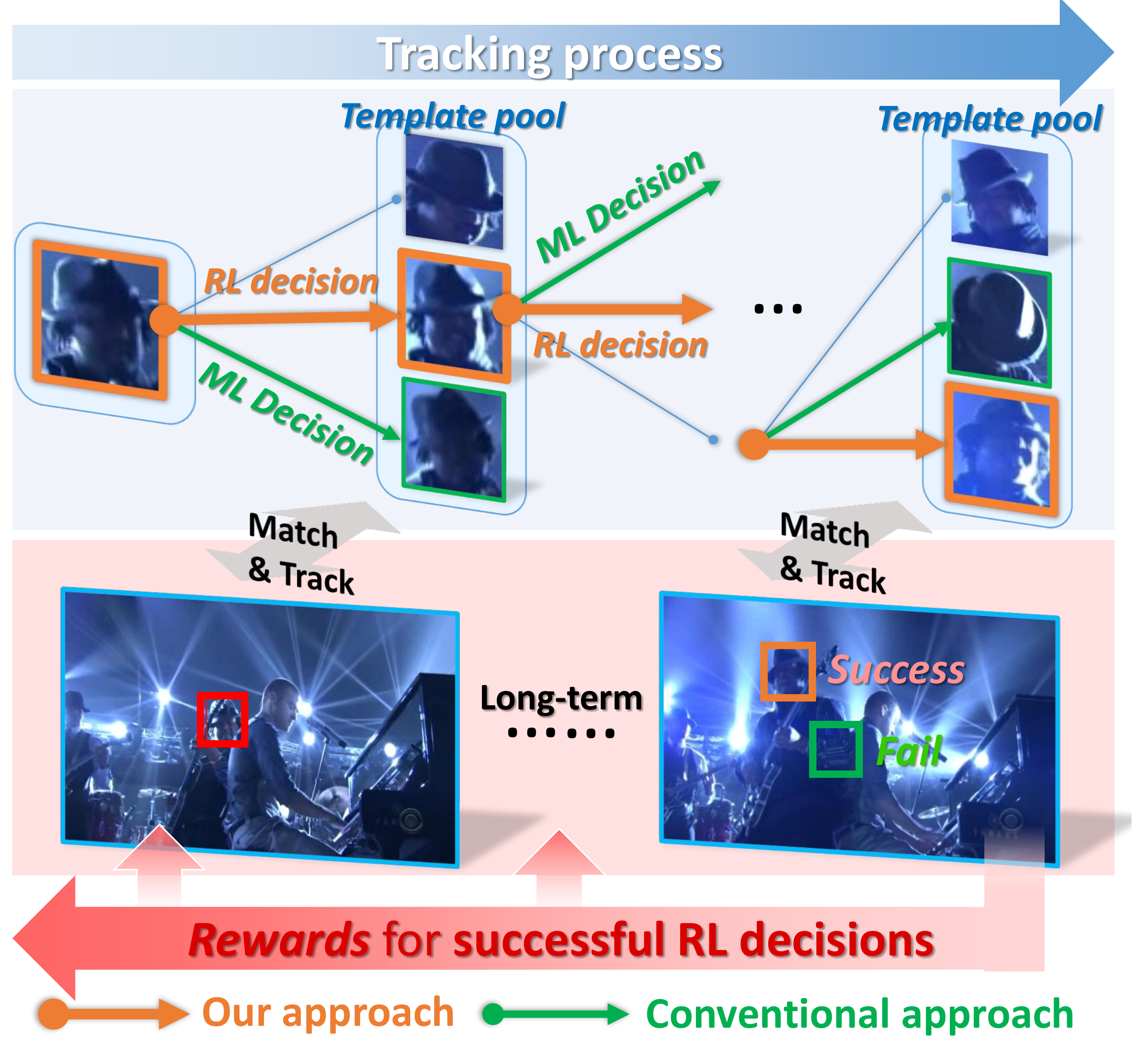}}
	\caption{\textbf{Motivation for the proposed visual tracking algorithm.} Our tracking algorithm formulates the visual tracking problem as a consecutive decision making task that can be self-learned through a reinforcement learning scheme. Rather than simply using the template with the maximum likelihood (ML decision), our algorithm strategically chooses the best template from the template pool in terms of localization and long-term success (RL decision).}
	\label{fig:1}
\end{figure}

\begin{figure*}
	\centering{\includegraphics[width=0.94\linewidth]{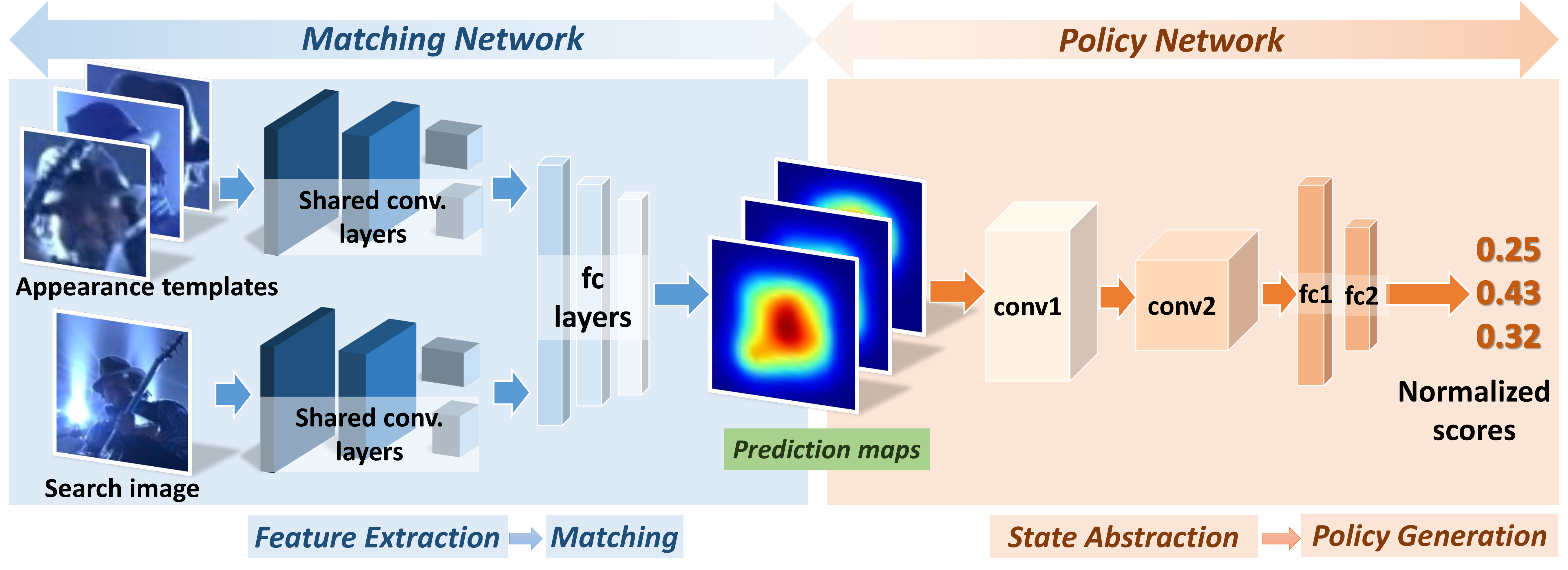}}
	\caption{\textbf{Overall architecture of the proposed system.} Matching network is a Siamese network which consists of shared convolutional layers as feature extractors and fully connected layers for matching. Matching result is passed to the policy network where it also consists of convolutional layers for state abstraction and fully connected layers for policy generation.}
	\label{fig:2}
\end{figure*}

Visual tracking is one of the most important and fundamental problems in the fields of computer vision and it has been utilized in many applications, such as automated surveillance, human computer interaction, and robotics. Also known as \textit{model-free} object tracking, visual tracking algorithms aim to track an arbitrary object throughout a video segment, given the object's initial location as a bounding box representation.

For several years, visual tracking problem has been regarded as a \textit{tracking-by-detection} problem, where the visual tracking task is formulated as an object detection task performed on cons`ecutive video frames. Tracking algorithm is often composed of a combination of appearance and motion models of the object. Especially, the appearance model is carefully designed to be robust to numerous appearance variations of the target object, where common challenges arise from changes in illumination, motion blur, deformation and occlusion from surrounding objects \cite{appearance}.

To solve the aforementioned challenges, two approaches are mainly utilized to cover the appearance variations of the target object. One approach is to update the appearance model of the target on-line in the tracking process \cite{TLD,VTD,IVT,CST,KCF,Struck}, gaining new examples on the way. This approach considers the visual tracking problem as a semi-supervised learning task where the initial sample is labeled while other samples are not. However, inaccurate and erroneous update often causes the tracker to fail and drift to the background \cite{template,meem}. The other approach is to utilize a feature representation scheme that is more robust to appearance perturbations while maintaining the discriminability between the target object and background objects \cite{nipstrack,stct,mdnet,fctrack,100fps}. This approach shares a common objective of other computer vision tasks such as object detection and semantic segmentation.

Recently, with growing attention on deep neural networks, especially convolutional neural networks (CNN) \cite{lenet}, there have been several approaches to utilize the powerful representation capabilities of CNNs for the visual tracking task. These methods showed successful results in covering the target appearance variations in short video segments. However, we also focus on the other aspect of visual tracking. Our proposed real-time visual tracking algorithm aims to utilize the deep neural network for revising the on-line update by making decisions concerning which template is the most adequate for localizing the target in a new frame. Our method formulates the visual tracking task as a consecutive decision making process where given past target appearance samples, the tracker has to decide which sample is the best for localizing the target for a new frame. Figure \ref{fig:1} illustrates the motivation of this research.

While there are large image datasets such as ImageNet \cite{imagenet} with ground truth labels available for obtaining a powerful feature representation under supervised learning environment, on-line update and selection of the target appearance model for visual tracking should be adequately tuned according to the tracking environment and the capacity of the feature representation that is used. This results in an absence of explicit labels on when and how to update the appearance model, which makes supervised learning infeasible. To resolve this problem, we adopt a reinforcement learning environment where given sequential states, an agent is prompted to make actions that can maximize the future reward. To achieve this learning task, we adopt deep neural networks for efficient state representation. Then we utilize policy gradient methods used in \cite{reinforce,asyncrl} and experience replay memory as in \cite{dqn-nips}, motivated by their recent success in playing the game of Go and ATARI video games in  \cite{alphago,dqn}. We train our policy network using randomly generated episodes from VOT-2015 tracking benchmark dataset \cite{vot}. We build our tracking algorithm based on a Siamese matching network by \cite{siamycnn} for its simplicity and real-time tracking speed while having a powerful representation capacity. To our knowledge, our work is one of the first to utilize a deep reinforcement learning methodology for on-line update in visual tracking. 

\begin{algorithm}[t]
	\SetKwData{Left}{left}\SetKwData{This}{this}\SetKwData{Up}{up}
	\SetKwFunction{Union}{Union}\SetKwFunction{FindCompress}{FindCompress}
	\SetKwInOut{Input}{input}\SetKwInOut{Output}{output}
	
	\Input{Tracking sequence of length $L$ \\ Initial target location $x_0$}
	\Output{Tracked target locations $x_t$} \BlankLine
	
	\emph{// For every frame in a sequence}\\
	\For{$t=1$ to $L$} {
		\emph{// For all N templates}\\
		\For{$i=1$ to $N$} {
			Produce prediction maps $s_i$ with each template $i$;\\
			Obtain normalized scores for each prediction map using policy network $\pi (a_i|s_i;\theta)$;\\
		}
		Find the prediction map $s_t$ with maximum score, corresponding template is chosen for localizing the target;\\
		Construct scale pyramid, obtain response maps and choose best scale with maximum value;\\
		Localize the target $x_t$ with shifted search images, choose mean position as target location $x_t$;\\
		Add a template to template pool every $K$ frames, discarding the oldest one;\\
		%		}
	}
	\caption{Visual tracking with reinforced decision making} \label{alg:test}
\end{algorithm}

\section{Related Work}

Conventional visual tracking algorithms can be largely categorized into two approaches. One approach constructs a generative model from previously observed samples and utilizes this model to find the region in the new frame where it can be described by the model best. The other uses a discriminative model where a classifier is learned to distinguish the target object region from the surrounding background region.

Generative approaches for visual tracking often utilize sparse representation as in \cite{MTT,CST,L1} or linear subspace for incremental learning as in \cite{VTD,IVT}. Using these criteria, they try to find the target region where it can be described by the model. The model is constructed from target appearance samples collected from previously tracked frames. \cite{IVT} uses principal component analysis on previous templates to construct a incremental subspace that can be used to reconstruct the target appearance. The target is localized by finding the location with the lowest reconstruction error. 
Discriminative approaches for visual tracking often utilize classifiers as in \cite{MIL,Struck,TLD} or correlation filters as in \cite{KCF,convcorr,Muster,color}. These approaches try to build a model that can distinguish the target appearance from the background region by using classification or regression. The model is trained from target and background appearance samples together. \cite{Struck} uses structured SVM to find the transformation vectors for patches obtained from the vicinity of the target, solving the label ambiguity problem of the binary classification assumption.
Other than the generative and discriminative methods, there are hybrid methods as in \cite{collab0,collab1} that aim to utilize the advantages of both models. \cite{collab1} adopted two components for the appearance model; with one descriptive and the other discriminative. Both components are integrated through a single optimization task.

Recently, there have been approaches to utilize deep representations for the visual tracking task. Convolutional neural networks (CNN) \cite{lenet} have shown outstanding performance in a wide range of computer vision applications including image classification \cite{alexnet}, object detection \cite{frcnn} and much more. Their powerful representation capacity motivated visual tracking approaches such as \cite{nipstrack,convcorr,stct,mdnet,fctrack}. \cite{nipstrack} was the first to introduce deep representation learning to visual tracking problem. They build a stacked denoising autoencoder and utilize its intermediate representation for visual tracking.
In \cite{convcorr}, hierarchical correlation filters learned on the feature maps of VGG-19 network \cite{vgg} are efficiently integrated. \cite{fctrack} also utilizes the feature maps generated from the VGG network to obtain multi-level information. \cite{mdnet} used the structure of low-level kernels of VGG-M network \cite{vgg} and trained on visual tracking datasets to obtain multi-domain representation for a robust target appearance model. 

Based on deep representations, some outstanding performances were shown by using two-flow Siamese networks on stereo matching problem in \cite{mccnn} and patch-based image matching problem in \cite{matchnet}. Accordingly, approaches to solve the visual tracking problem as a patch matching problem have emerged in \cite{siamis,siamfc,siamycnn,100fps}. \cite{siamis} and \cite{100fps} train the Siamese networks using videos to learn a patch similarity matching function that shares an invariant representation. \cite{siamfc} and \cite{siamycnn} further expand this notion and proposes a more end-to-end approach to similarity matching where a Siamese architecture can localize an exemplar patch inside a search image using shared convolutional layers. In particular, \cite{siamfc} proposes a fully-convolutional architecture that adopts a cross-correlation layer to obtain invariance to spatial transitions inside the search image, lowering the complexity of the training process significantly.

However, approaches such as \cite{100fps,siamycnn} use a naive on-line update strategy that cannot revise erroneous updates and recover from heavy occlusions. Moreover, approaches \cite{siamis,siamfc} do not update the initial template, solely relying on the representation power of the pre-trained CNN. This approach may be effective for short-term video segments with no distractors, but the tracker can be attracted towards a distractor with a similar appearance to the target. Our proposed algorithm is aimed to solve both problems of the previous approaches, by utilizing previously seen examples to adapt to the recent appearance of the target and choosing the most adequate template for localizing the target, ruling out erroneously updated templates. 

There also have been some recent approaches that employ deep reinforcement learning methodology on visual tracking algorithms. \cite{adnet} trained a policy network to generate actions for state transition in order to localize the target in a given frame. \cite{ptrack} used YouTube videos to interactively learn a Q-value function where it makes decisions for the tracker to reinitialize, update or to keep tracking with the same appearance model. However, both trackers run at 3 fps and 10 fps respectively, both lacking the speed of real-time performance. Our algorithm runs at a real-time speed of 43 fps while maintaining a competitive performance. We achieve this by incorporating more lightweight and optimized structures for matching network and policy network.

\begin{algorithm}[t]
	\SetKwData{Left}{left}\SetKwData{This}{this}\SetKwData{Up}{up}
	\SetKwFunction{Union}{Union}\SetKwFunction{FindCompress}{FindCompress}
	\SetKwInOut{Input}{input}\SetKwInOut{Output}{output}
	
	\Input{Randomly generated episode of length $L$, Policy network weights $\theta^{-}$}
	\Output{Updated policy network weights $\theta^{+}$} \BlankLine
	
	\emph{// For every frame in an episode}\\
	\For{$t=1$ to $L$} {
		\emph{// For all N templates}\\
		\For{$i=1$ to $N$} {
			Produce prediction maps $s_t$ with each template $i$;\\
			Obtain normalized scores for each prediction map using policy network $\pi (a_i|s_t;\theta^{-})$;\\
		}
		Choose some $a_t \in \{a_i,...,a_N\}$ stochastically, with probabilities proportional to the normalized scores;\\
		Obtain update gradient $\nabla_\theta \log \pi(a_t|s_t;\theta^{-})$;\\
		Accumulate gradients according to eq. \ref{eq:3}, obtain $\Delta \theta$;\\
		Localize the target position $x_t$ as in Alg. \ref{alg:test};\\
		Add a template to template pool every $K$ frames, discarding an oldest template;\\
	}
	
	\eIf{episode successful}{
		Update weights $\theta^{+} = \theta^{-} + \Delta \theta$;\\
	}{Update weights $\theta^{+} = \theta^{-} - \Delta \theta$;\\}
	Obtain $4N$ samples from experience replay memory, calculate gradients and add to $\theta^{+}$;\\
	\caption{Training the policy network for a single episode} \label{alg:train}
\end{algorithm} \DecMargin{1em}

%%%%%%%%% PROPOSED ALGORITHM %%%%%%%%%%%%
\section{Proposed Algorithm}

In the following subsections, we first show a brief overview of our proposed visual tracking algorithm. Then we describe the details of the proposed method. We show the theoretical background for the reinforcement learning formulation. Next we describe the architectures for visual tracking and its training scheme.

\subsection{Tracking with Reinforced Decisions}

Our tracking system can be divided into two parts where the first part is the matching network that produces prediction heatmaps as a result of localizing the target templates inside a given search image. And the second part is the policy network that produces the normalized scores of prediction maps obtained from the matching network. Figure \ref{fig:2} shows the overall diagram of our tracking system.

Assuming the networks are trained and its weights are fixed, we can perform the visual tracking task on arbitrary sequences. For a given video frame, we crop and obtain a search image based on the target's previous bounding box information. Search image is cropped with its center location being the target's previous center location and its scale being double the target's previous scale. Using the appearance templates obtained from previously tracked frames, the matching network produces prediction maps for each appearance template. Then each of the prediction maps are fed to the policy network where it produces the normalized scores for each prediction map. The prediction map with the maximum score is chosen and its corresponding template is used to track and localize the target. The position in the prediction map with the maximum value corresponds to the next position of the target in the search image. To obtain a more precise scale and position of the target, we construct a scale pyramid of 3 images using the search image to adapt to the scale change of the target. Scale with the maximum response value is chosen. Then we additionally use 4 shifted versions of the search image to obtain a more fine-level position of the target. Size of the $x,y$-axis shift are fixed proportionally to the target width/height respectively. Maximum positions in each response map are found and we calculate the mean position as the final location of the target.

Appearance templates are obtained on a regular time interval during the tracking process. We intentionally use this simplistic method for updating the template pool in order to promote robustness in the decision making process. This encourages the policy network to be trained on various situations where the policy network is expected to avoid choosing an outlier template (i.e. dark, blurry, occluded template) which may be present in the template pool. The overall flow of our tracking algorithm is described in algorithm \ref{alg:test}.

%Using the trained policy network, we can perform the visual tracking task on arbitrary sequences. The tracking process is not much different from the training process as illustrated in algorithm \ref{1}. At test time, the policy network weights are fixed and we select actions in a deterministic way by choosing the template with the best score. We intentionally use a simplistic template update strategy to promote robustness in decision making. For practical reasons, we average the predictions obtained from 5 slightly shifted version of search images to obtain a finer prediction. And to cover the scale space, we use 3 scaled versions of the search image to find the best scale that fits the template.

\subsection{Reinforcement Learning Overview and Application to Visual Tracking}

%\begin{figure}[t]
%	\centering{\includegraphics[width=0.80\linewidth]{fig_pg.pdf}}
%	\caption{A conceptual visualization of policy refinement process.}
%	\label{fig:pg}
%\end{figure}

We consider a general reinforcement learning setting where the agent interacts with an environment through sequential states, actions and rewards. Given an environment $\mathcal{E}$ and its representative state $s_t$ at time step $t$, an agent must perform an action $a_t$ selected from a set $\mathcal{A}$ of every possible actions. Action $a_t$ is determined by the policy $\pi(a_t|s_t)$ where action $a_t$ can be chosen with deterministic or stochastic manner. In return for the action, the agent receives a scalar reward $r_t$ and observes the next state $s_{t+1}$. This recurrent process continues until the agent reaches a terminal state. The goal of the agent is to select actions that maximizes the discounted sum of expected future rewards, where we define the action-value function as $Q^{\pi} (s,a)$ \cite{rlbook}.

%\begin{equation}\label{eq:1}
%Q^{\pi} (s,a) = \mathbb{E}[ r_t + \gamma r_{t+1} + \gamma^2 r_{t+2} + ... | s_t=s, a_t=a, \pi ]
%\end{equation}
%where $\gamma \in (0,1]$ is a discount factor to ensure convergence of the accumulated return.

To achieve the goal mentioned above, there are mainly two approaches to reinforcement learning; value-based methods and policy-based methods. Value-based method assumes that there exists an optimal action-value function $Q^*(s,a)=\max_\pi Q^\pi (s,a)$ that gives the maximum action-value for a state-action pair, given some implicit policy (e.g. $\epsilon$-greedy) \cite{qlearn}.
%The aim of value-based methods is to approximate the optimal action-value function using a function approximator as in $Q^*(s,a){\approx}Q(s,a;\theta)$ where $\theta$ denotes the parameter for the function approximator. Minimization of the discrepancy between two functions can be achieved through variants of Q-learning algorithms \cite{dqn}.

\begin{figure}[t]
	\centering{\includegraphics[width=1.00\linewidth]{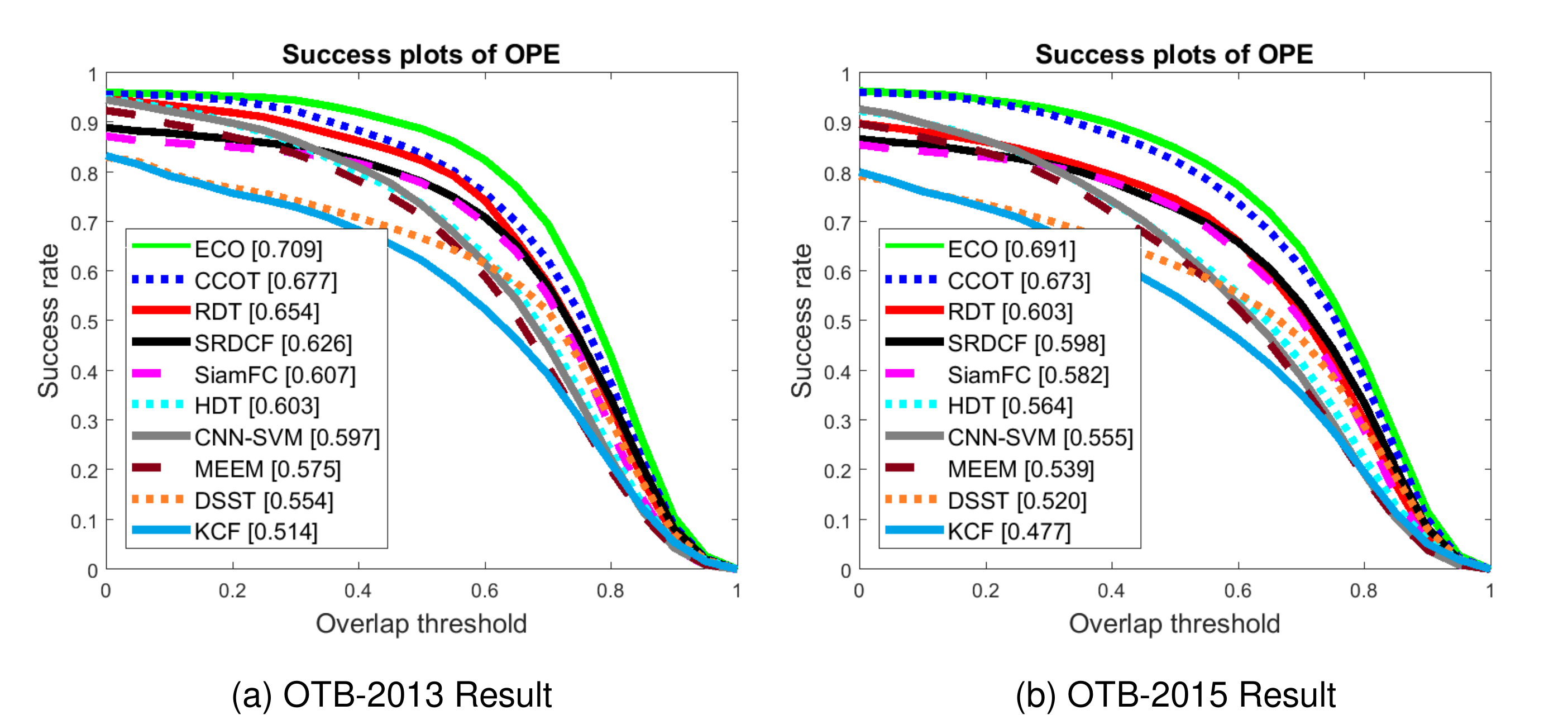}}
	\caption{OPE result comparison on (a) OTB-2013 and (b) OTB-2015 benchmark dataset \cite{otb}. The numbers in the legend box indicate the average area-under-curve (AUC) scores for each tracker.}
	\label{fig:3}
\end{figure}

On the other hand, policy-based methods \cite{pgm} aim to directly model the policy function $\pi(a|s;\theta)$ without the assumption of intermediate action-value function, removing the need of evaluating the values of possible actions for a given state. Approximation of the policy function can be achieved by maximizing the objective return function $R_t$ using stochastic gradient ascent algorithms. One simplest example of the method is the REINFORCE algorithm introduced in \cite{reinforce} where gradient ascent is used on expected reward $R_t=\mathbb{E}[r_t]$ as in

\begin{equation}\label{eq:2}
\Delta \theta = \alpha \nabla_\theta \log \pi(a_t|s_t;\theta) r_t ,
\end{equation}

where $\alpha$ is the learning rate. %Figure \ref{fig:pg} shows a conceptual explanation for the policy refinement\footnote{Explanatory figure inspired by Andrej Karpathy's article, \url{http://karpathy.github.io/2016/05/31/rl/}}.% 
First, actions are sampled from the policy distribution $\pi$, then the performed actions are evaluated and rewards are given from interacting with the environment. Using this information, we can refine the policy through a parameter update. Samples from an updated policy is expected to give us a higher reward. For our work, we use a variant of policy-based reinforcement learning method since it is commonly known to have better convergence properties and capability of learning stochastic policies \cite{asyncrl}.

To apply this reinforcement learning method to our visual tracking environment, we define state $s_t$ of the tracker as the overall combination of prediction maps which are obtained from the matching network using the corresponding appearance templates gained in the course of tracking. The action $a_t$ is defined as selecting a single template from current template pool to locate the target in a given frame. 
This can be simulated by drawing a sample from a discrete probability distribution of normalized scores where each score is produced by the policy network using the corresponding prediction map. At training time, we can train the policy network to assign higher scores to prediction maps (templates) that result in successful tracking using a rewarding scheme.

Reward $r_t$ is given at the end of a tracking episode depending on success or failure of a tracking episode. Positive reward will be given when the tracker successfully tracks the target, producing a bounding box overlap score over a predefined threshold. This will encourage the policy network to perform more actions (choosing templates) that may result in a successful tracking episode. On the other hand, negative reward will be given when the tracker loses the target as a result of performing a chain of poor actions. This will help the tracker to avoid performing those actions in the future, thus resulting in more successful tracking episodes hereafter. At test time, we select the prediction map (and its corresponding appearance template) with the highest normalized score to locate the target in a given frame. 

\begin{figure}[t]
	\centering{\includegraphics[width=1.00\linewidth]{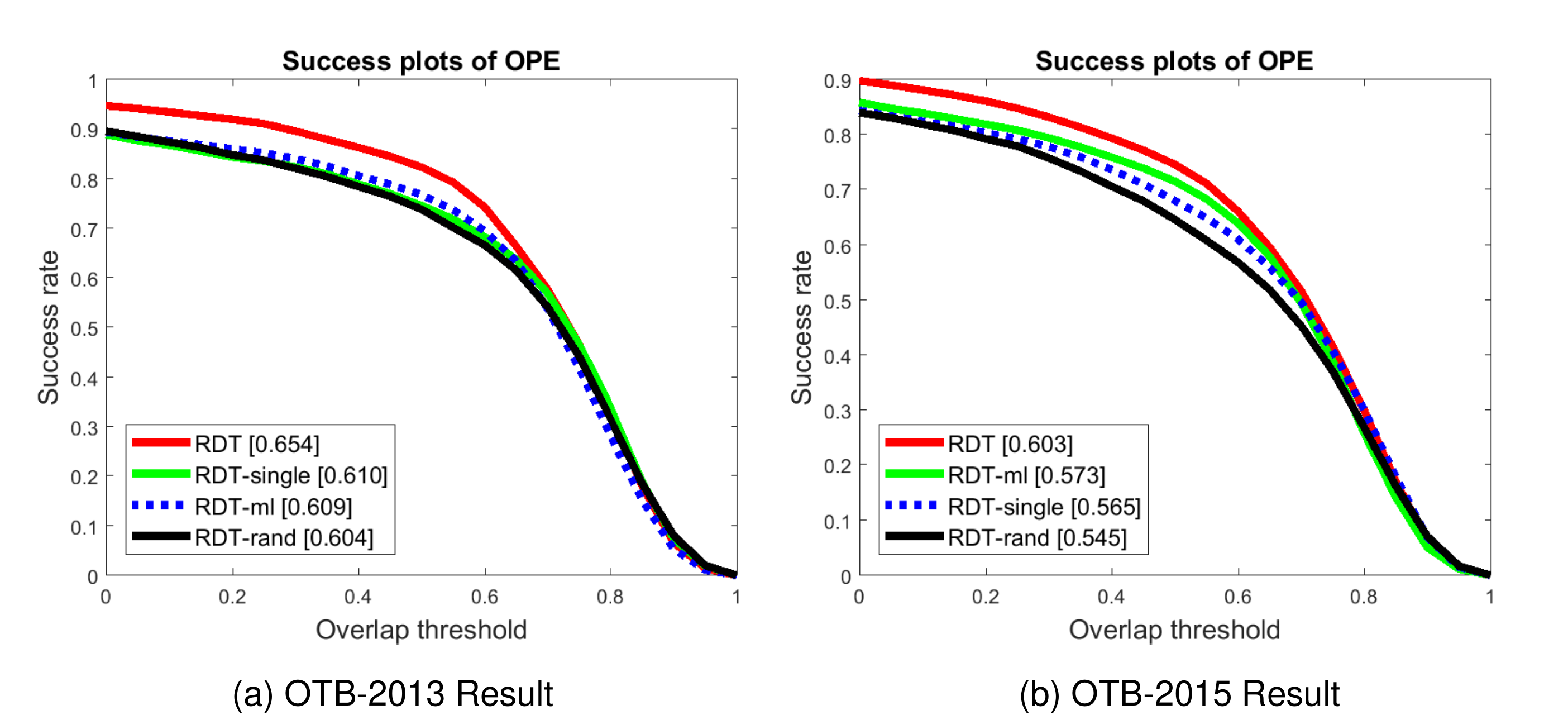}}
	\caption{OPE result comparisons of (a) OTB-2013 and (b) OTB-2015 benchmark dataset \cite{otb} for the internal comparisons.}
	\label{fig:4}
\end{figure}
\subsection{Network Architectures}

\begin{table}[t]
	\centering
	\begin{small}
		
		\begin{tabular}{c c c c c}
			\hline
			& \textbf{RDT} & \textbf{RDT-ml} & \textbf{RDT-rand} & \textbf{RDT-single} \\
			\hline
			\textbf{SRE} & 0.601 & 0.565 & 0.544 & 0.543 \\
			
			\textbf{TRE} & 0.646 & 0.625 & 0.605 & 0.615 \\
			\hline
		\end{tabular}
		
	\end{small}
	\label{table:mpe}
	\caption{SRE and TRE tracking performances measured in AUC on OTB-2015 benchmark dataset for internal comparisons.}
\end{table}

\textbf{Architecture of the Matching Network:} We borrow the Siamese architecture design from \cite{siamycnn}, using it as our baseline tracker. The Siamese architecture has 2 input branches and uses 3 shared convolutional layers for extracting the common representations from the object patch and the search patch. Then these features are concatenated and fed to 3 fully connected layers to produce a Gaussian prediction map, where the maximum point is the relative location of the object patch inside the search patch.

\begin{figure*}[t]
	\centering{\includegraphics[width=0.95\linewidth]{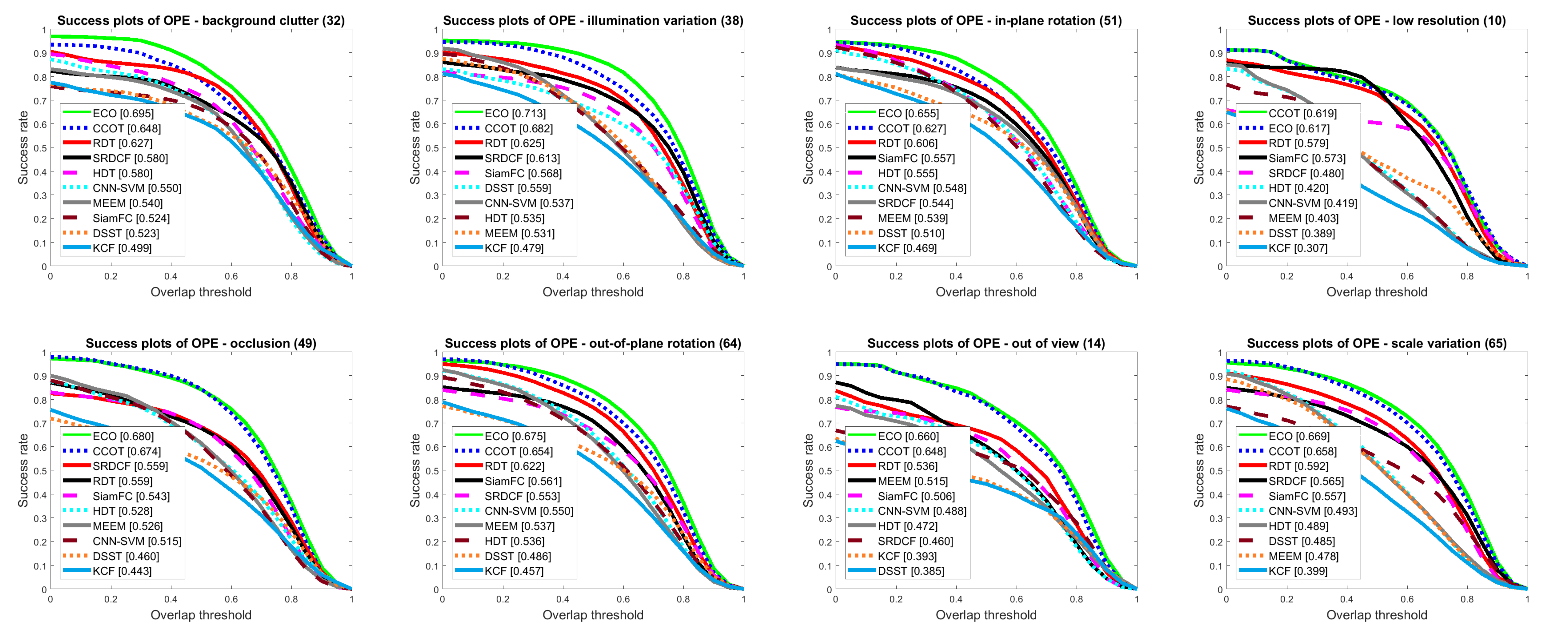}}
	\caption{Success plots for 8 challenge attributes: background clutter, illumination variation, in-plane rotation, low resolution, occlusion, out-of-plane rotation, out of view and scale variation.}
	\label{fig:attrib}
\end{figure*}

\textbf{Architecture of the Policy Network:} Our policy network sees the output prediction maps produced by the matching network and makes the decision whether the matching result is reliable or not. By following this decision, we can always choose the most appropriate template for locating the target object in a given frame. The policy network consists of 2 convolutional layers to produce an adequate representation of the state and 2 fully connected layers for deciding whether this state is reliable or not. Then the outputs are fed through sigmoid function to produce probabilities. Finally, we choose the activation with the highest value and its corresponding template as the best candidate for tracking.

\begin{table}
	\centering{
		\small{
			\resizebox{0.80\linewidth}{!}{
				\begin{tabular}{|m{0.7\linewidth}|c|}
					\hline
					\multicolumn{2}{|c|}{input ($48\times48$, $120\times120$ RGB image)}\\
					\hline \hline
					\multicolumn{2}{|c|}{conv7,3-16}\\ \hline
					\multicolumn{2}{|c|}{pool2,2, relu}\\ \hline
					
					\centering{conv3,1-32} & \multirow{4}{0.3\linewidth}{\centering{conv1,1-4}}\\
					
					\cline{1-1}
					\centering{pool2,2, relu}&\\
					\cline{1-1}
					\centering{conv3,1-64}&\\
					\cline{1-1}
					\centering{pool2,2}&\\ \hline
					
					\multicolumn{2}{|c|}{relu, reshape, concat}\\ \hline
					\multicolumn{2}{|c|}{fc-2048, relu}\\ \hline
					\multicolumn{2}{|c|}{fc-2048, relu}\\ \hline
					\multicolumn{2}{|c|}{fc-961}\\ \hline
					\multicolumn{2}{|c|}{sigmoid, reshape}\\ \hline \hline
					\multicolumn{2}{|c|}{output ($31\times31$, prediction map)}\\			
					\hline
				\end{tabular}
			}
		}
	}
	\caption{Architecture of the matching network. The convolutional layer parameters are denoted as conv(kernel size, stride)-(number of channels) and fully conneced layer parameters are denoted as fc-(number of units). Max pooling layer is denoted as pool-(kernel size, stride)}
	\label{table:ycnn}
\end{table}

\subsection{Training the Policy Network}

\begin{table}
	\centering{
		\small{
			\begin{tiny}
			\resizebox{0.80\linewidth}{!}{
				\begin{tabular}{l c c}

					\hline
					\textbf{Tracker} & \textbf{AUC} & \textbf{FPS}\\
					\hline
					ECO \cite{ECO} & 0.691 & 8\\
					\textbf{RDT} (Ours) & \textbf{0.603} & \textbf{43}\\
					SRDCF \cite{srdcf} & 0.598 & 5\\
					CSR-DCF \cite{csrdcf} & 0.587 & 13\\
					SiamFC \cite{siamfc} & 0.582 & 58 \\
					CFNet-conv2 \cite{cfnet} & 0.568 & 75\\
					HDT \cite{hdt} & 0.564 & 10\\
					DSST \cite{dsst} & 0.520 & 24 \\
					KCF \cite{KCF} & 0.477 & 172\\				
					GOTURN \cite{100fps} & 0.427 & 100\\
					TLD \cite{TLD} & 0.406 & 20 \\
					\hline
					
				\end{tabular}
			}
			\end{tiny}
		}
	}
	\caption{Tracking performance on OTB-2015 dataset and speed comparison between trackers.}
	\label{table:fps}
\end{table}

To train the policy network $\pi (a|s;\theta)$ introduced above, we use a variant of REINFORCE algorithm with accumulated policy gradients. We randomly generate numerous tracking episodes with varying lengths from VOT-2015 \cite{vot} video dataset. Then we perform tracking on each training episodes with stochastically sampled action roll-outs produced by the policy network to ensure exploration of state space. For each decisions in an episode, we temporarily assume each decision was optimal and perform backpropagation to obtain gradients for all weights inside the policy network. We accumulate these gradients for all decisions in a single episode as in

\begin{equation}\label{eq:3}
\Delta \theta = \alpha \sum_{t=1}^{L} \nabla_\theta \log \pi(a_t|s_t;\theta) \beta^{L-t} ,
\end{equation}
where $L$ is the length of an episode, $\alpha$ is the learning rate and $\beta \in (0,1]$  is a discounting factor inserted to give more weight to decisions made later in the episode. If an episode terminates, weights in the policy network are updated according to the success or failure of that episode. If an episode was successful, gradient is updated accordingly. If an episode was a failure, negative gradient is applied. The overall algorithm for training the policy network for a single episode is described in algorithm \ref{alg:train}.

We also keep an experience replay memory of state-action-reward gained from previous episodes. When gradients are applied to the policy network after an episode, experiences are randomly sampled from the experience replay memory and applied concurrently. Successful experiences and failure experiences are kept separately in each replay memories. At every training step, total update gradient $\Delta\theta$ is a sum of $5L$ gradients where $L$ gradients are obtained by decisions from a single episode and $4L$ gradients are sampled from the decisions in the experience memory ($2L$ from each success and failure memory). By using these sampled experiences from the experience replay memory, we can remove the correlation in incoming data sequence and reduce the variance of the update to obtain more stable convergence.

%%%%%%%%% EXPERIMENTS %%%%%%%%%%%%
\section{Experiments}

\subsection{Implementation Details}
\begin{figure}[t]
	\centering{\includegraphics[width=0.95\linewidth]{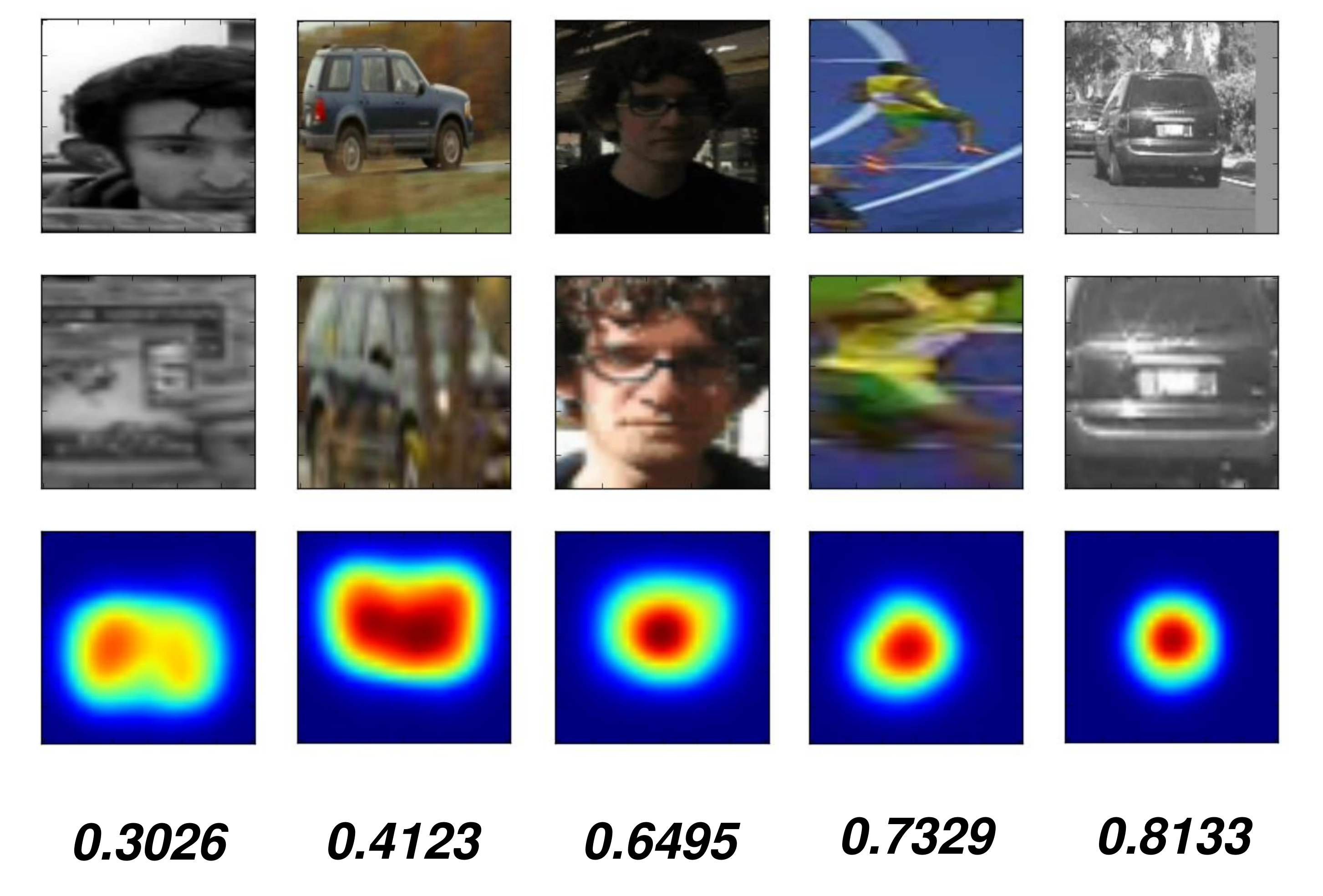}}
	\caption{Example input search image (1st row), template image (2nd row), prediction maps (3rd row) and their corresponding output scores of the policy network (4th row). Outputs are shown in values [0,1]}
	\label{fig:policy}
\end{figure}

\textbf{Matching network parameters: } Input to the matching network are $48\times48$ appearance template and $120\times120$ search image. Matching network consists of the three convolutional layers with a skip connection layer connecting the first and the third layer. Activations are then fed to the 3 fully connected layers with dropout rate of $0.8$, followed by a sigmoid function. Output is a $31\times31$ prediction map. More detailed network parameters are shown in table \ref{table:ycnn}.

\textbf{Policy network parameters: } Input to the policy network are $31\times31$ prediction maps. The first convolutional layer has a $5\times5$ sized kernel with 4 output channels and it is applied with a stride of 3, then the activations are $2\times2$ max-pooled. The second convolutional layer has a $3\times3$ sized kernel with 8 output channels and it is applied with a stride of 1. Then the activations are fed to fully-connected layers, each with 128 hidden activations and 1 activation. For fully-connected layers, dropout regularization \cite{dropout} is used with keep probability of $0.7$. All layers are initialized from Gaussian normal distribution with zero mean and variance of $0.1$ and each convolutional layer is followed by rectified linear unit (ReLU) activation functions. 

\textbf{Training parameters: } To train the matching network, batch size of 64 is sampled from the ImageNet \cite{imagenet} dataset. For optimization, Adam optimizer \cite{adam} with learning rate of $10^{-4}$ is used. For the policy network, Adagrad optimizer \cite{adagrad} with learning rate of $10^{-4}$ is used and $\beta=0.95$ is used. We train our policy network using 50,000 episodes randomly sampled from the VOT-2015 \cite{vot} benchmark dataset. Overlapping sequences with OTB dataset were removed before training. Length of each episode is between 30 and 300 frames and a new template is added to the template pool every 50 frames. Success or failure of an episode is determined by the mean intersection-over-union (IoU) ratio of last 20 bounding box predictions compared to the ground truth bounding boxes. If the mean IoU is under 0.2, we consider the episode as a failure. To lower the variance of each update, we keep an experience replay memory for 5000 successful samples and 5000 failure samples. Setting the episode length too short will result a deficiency in number of failure experiences to learn from. For each update, 40 samples are sampled from each experience replay memory and gradient is applied concurrently. 

\begin{table}[t]
	\setlength{\tabcolsep}{0.3em}
	\centering
	\begin{small}
		\begin{tabular}{c | c c c c c c c c}
			\hline
			\textbf{Frames} & 20 & 40 & 50 & 80 & 100 & 150 & 200 & 300\\
			\hline
			\textbf{AUC} & 0.615 & 0.605 & 0.603 & 0.595 & 0.599 & 0.596 & 0.587 & 0.586 \\
			\hline
		\end{tabular}
	\end{small}
	\caption{Tracking performance on OTB-2015 benchmark dataset under different template update intervals (frames).}
	\label{table:interval}
\end{table}	

\textbf{Tracking parameters: } For practical reasons, we also average the location predictions obtained from 4 slightly shifted (upward, downward, left and right) search images to obtain a more accurate localization. Shift sizes are fixed to 20\% of the target's width/height. Additionally, to cover the scale space, we use 3 scaled versions of the search image to find the best scale that fits the template. Scale parameters used are $1.05, 1.00$ and $1.05^{-1}$. We used a maximum of 4 templates including the initial template for tracking, and template pool is updated every 50 frames, replacing the oldest template. Increasing the number of maximum templates beyond this limit did not show noticeable performance gain since the tracker tended to utilize a small subset of templates that were updated more recently. This tendency is coherent with the results shown in \cite{meem} where number of actively utilized snapshots was limited. \textit{All parameters were fixed} throughout the evaluation of the benchmark dataset.

\begin{figure*}[t]
	\centering{\includegraphics[width=0.90\linewidth]{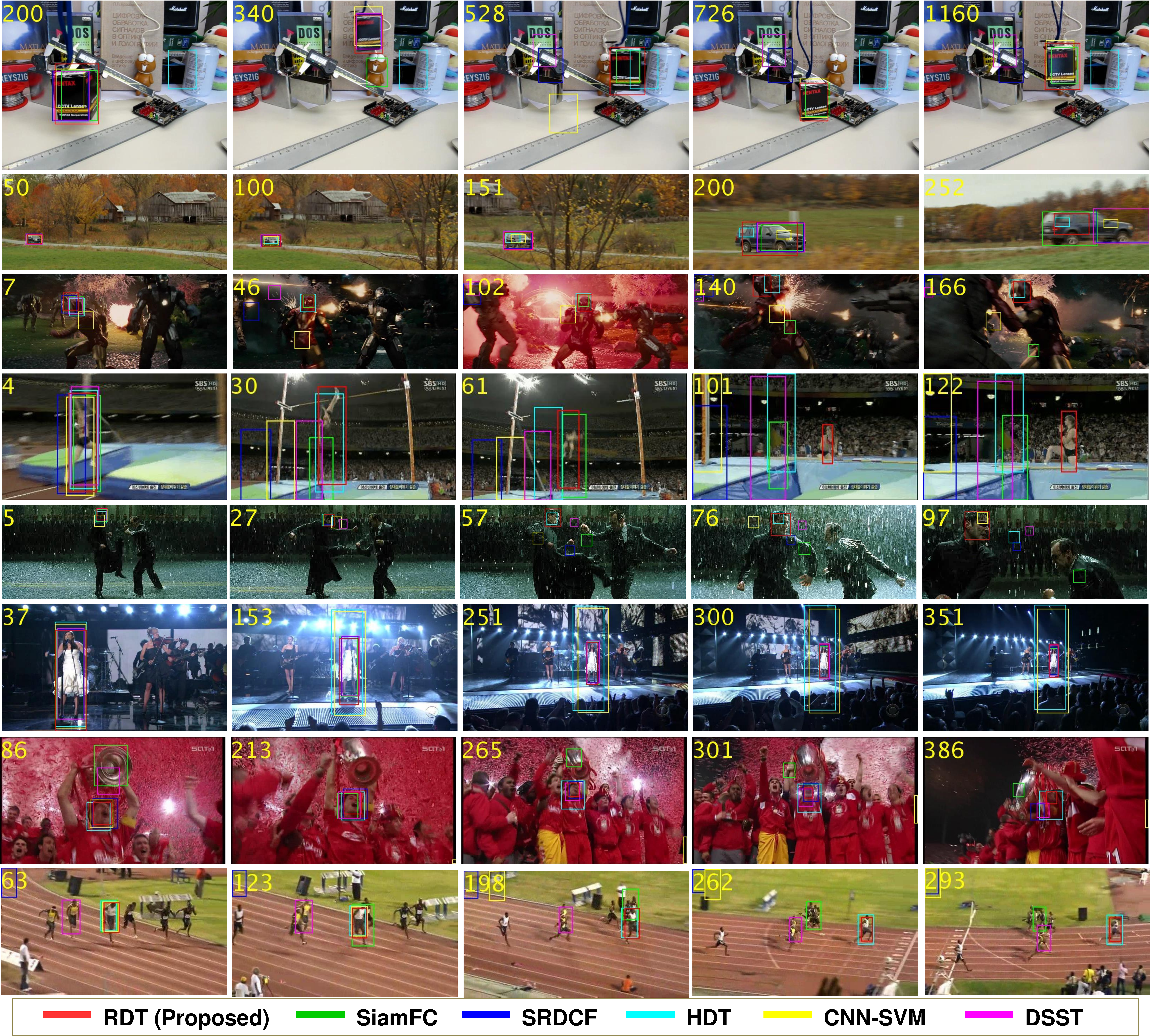}}
	\caption{Qualitative results of the proposed method on challenging sequences from OTB benchmark dataset (in vertical order, \textit{box, carScale, ironman, jump, matrix, singer1, soccer} and \textit{bolt2})}
	\label{fig:qual}
\end{figure*}

\textbf{Implementation environment: } We implement our tracker in Python using TensorFlow \cite{tensorflow} library. The implementation runs on an Intel Core i7-4790K 4GHz CPU with 24GB of RAM and the neural network is computed and trained on GeForce GTX TITAN X GPU with 12GB of VRAM. Our implemented tracker runs at an average of 43 frames per second (FPS) on OTB-2015 \cite{otb} video dataset.

\subsection{Evaluation on OTB dataset}

\subsubsection{Quantitative Results}

Object tracking benchmark (OTB-2015) \cite{otb} is a well-known visual tracking benchmark dataset that contains a total of 100 fully annotated sequences. OTB-2013 is a subset of the OTB dataset which contains 50 selected sequences from the original dataset.  We compare our tracking algorithm with 9 other tracking algorithms, including 5 real-time algorithms. SiamFC \cite{siamfc}, SRDCF \cite{srdcf}, HDT \cite{hdt}, CNN-SVM \cite{cnnsvm}, MEEM \cite{meem}, DSST \cite{dsst}, KCF \cite{KCF}, ECO \cite{ECO} and CCOT \cite{CCOT} tracking algorithms are used for comparison. Success plots for both OTB-2013 and OTB-2015 sequences are shown on figure \ref{fig:3} where the proposed algorithm is denoted as RDT. Success rate evaluation metric is calculated by comparing the predicted bounding boxes with the ground truth bounding boxes to obtain the IoU scores and measuring the proportion of scores larger than a given threshold value. Final score is calculated by measuring the area-under-curve (AUC) for each tracker. The result shows that our proposed algorithm achieves a competitive performance alongside with other trackers, despite using a simple patch matching framework. Our tracker also performs at a real-time speed of 43 FPS with relatively high performance (Table \ref{table:fps}), compared to other deep representation based trackers such as HDT \cite{hdt} running at 10 FPS, SRDCF \cite{srdcf} running at 5 FPS and ECO \cite{ECO} running at 8 FPS.. The real-time processing speed of our tracker is a favorable characteristic for training the policy network using numerous training episodes.

We further analyze the performance of our tracker for 8 different challenge attributes labeled for each sequence, where sequences with background clutter, illumination variation, in-plane rotation, low resolution, occlusion, out-of-plane rotation, out of view and scale variation are evaluated. As shown in figure \ref{fig:attrib}, our tracker shows competitive results on most of the attributes compared to the other trackers. 

To show the effectiveness of our reinforced template selection strategy, we also perform internal comparisons between different baselines of our algorithm. Figure \ref{fig:4} shows the OPE plots and Table \ref{table:mpe} shows the SRE and TRE results between four algorithms where RDT is the original algorithm, RDT-rand is a variant where the template is selected at random, RDT-ml is a variant where the template with the maximum likelihood score is selected and RDT-single is a variant where only the original template is used with no update. We were able to obtain a OPE performance gain of roughly 10\% for both OTB-2013 and OTB-2015 sequences by using the proposed template selection strategy, proving that our policy network chooses the more adequate template for tracking a given frame. 

The performance gain in TRE is smaller in comparison to OPE and SRE results. However, by the nature of TRE evaluation where an original OTB sequence is divided into 20 uniform time-segments and tested 20 times with initialization from each segment, overall performance is measured with sequences that are much shorter than their original sequences. Since our algorithm's performance gain comes from the appropriate update strategy in long-term sequences (i.e. \texttt{blurFace, basketball, panda, human6, faceOcc1, girl2, blurCar1, doll, ...}), smaller performance gain in short-term sequences is reasonable.

We also perform an additional experiment with different update intervals (frames) where the results are shown in Table \ref{table:interval}. Performance is measured by AUC tested on OTB-2015 dataset. Decreasing the update interval gives some performance gain while increasing the interval results loss in performance. We speculate that templates that are more recent are effective in localizing the target more precisely, while older templates are less effective and thus less utilized. This is consistent with our experiment with increased template pool where increasing the template over certain limit did not show noticeable performance gain.

\subsubsection{Qualitative Results}
Figure \ref{fig:qual} shows the snapshots of tracking results produced by the proposed algorithm with SiamFC \cite{siamfc}, SRDCF \cite{srdcf}, HDT \cite{hdt}, CNN-SVM \cite{cnnsvm} and DSST \cite{dsst}. Trackers were tested on some challenging OTB-2015 sequences (\textit{box, carScale, ironman, jump, matrix, singer1, soccer} and \textit{bolt2}) where selected frame numbers are denoted in yellow on the top-left corners respectively. Our proposed tracking algorithm performs robustly, without losing track of the target under challenging conditions such as occlusion in \textit{box} and \textit{soccer}, scale change in \textit{carScale}, \textit{singer1} and \textit{jump}, illumination variation in \textit{ironman} and \textit{matrix}. From the qualitative results, it is shown that our tracker successfully utilizes both the deep representation power and the template selection strategy for tracking the target.

We also show some example input/output pairs for the policy network in figure \ref{fig:policy} to show what the policy network has learned from numerous tracking episodes. Interestingly, the policy network has self-learned to avoid ambiguous decisions with high uncertainty (left) while preferring decisions with low ambiguity (right) by assigning a higher score on the prediction map.  Our tracker avoids utilizing erroneously updated templates (e.g. occluded samples) by choosing the best template in terms of uncertainty, resulting a more successful tracking episode.

%%%%%%%%% CONCLUSIONS %%%%%%%%%
\section{Conclusion}

In this paper, we proposed a novel tracking algorithm based on a template selection strategy constructed by deep reinforcement learning methods, especially policy gradient methods. Our goal was to construct a policy network that can choose the appropriate template from a template pool for tracking an arbitrary frame where the policy network is trained from numerous training episodes randomly generated from a tracking benchmark dataset. Experimental results show that we achieved a noteworthy performance gain in tracking under challenging scenarios, proving that our learned policy effectively chooses the appearance template that is more appropriate for a given tracking scenario. Our algorithm also performs at a real-time speed of 43 fps while maintaining a competitive performance compared to other real-time visual tracking algorithms.

\section*{Acknowledgments}
This work was supported by the Visual Turing Test project (IITP-2017-0-01780) from the Ministry of Science and ICT of Korea.

%-------------------------------------------------------------------------

%\begin{figure*}
%\begin{center}
%\fbox{\rule{0pt}{2in} \rule{.9\linewidth}{0pt}}
%\end{center}
%   \caption{Example of a short caption, which should be centered.}
%\label{fig:short}
%\end{figure*}
%
%\begin{table}
%\begin{center}
%\begin{tabular}{|l|c|}
%\hline
%Method & Frobnability \\
%\hline\hline
%Theirs & Frumpy \\
%Yours & Frobbly \\
%Ours & Makes one's heart Frob\\
%\hline
%\end{tabular}
%\end{center}
%\caption{Results.   Ours is better.}
%\end{table}

{\small
\bibliographystyle{ieee}

}

\end{document}